\pdfoutput=1

\documentclass[11pt]{article}

\usepackage{ACL2023}

\usepackage{times}
\usepackage{latexsym}

\usepackage[T1]{fontenc}

\usepackage[utf8]{inputenc}

\usepackage{microtype}

\usepackage{inconsolata}

\usepackage{enumitem}
\usepackage{tabularx}
\usepackage{tikz}
\usepackage{pgfplots}
\usepackage{caption}
\usepackage{subcaption}
\usepackage{graphicx}
\usepackage{MnSymbol}
\usepackage{makecell}
\usepackage{multirow}
\usepackage{booktabs}
\usepackage{arydshln}
\usepackage{CJKutf8}
\usepackage{bbm}
\usepackage{pifont}
\usepackage{amsmath} 
\usepackage{CJKutf8}

\title{Fine-Grained Self-Endorsement Improves Factuality and Reasoning}

\author{Ante Wang$^{1,2}$, Linfeng Song$^{3}$, Baolin Peng$^{3}$, Ye Tian$^{3}$, Lifeng Jin$^{3}$, Haitao Mi$^{3}$,\\
{\bf Jinsong Su}$^{1,2}$ and {\bf Dong Yu}$^{3}$ \\
$^{1}$School of Informatics, Xiamen University, China \\
$^{2}$Key Laboratory of Digital Protection and Intelligent Processing of Intangible Cultural Heritage\\of Fujian and Taiwan (Xiamen University), Ministry of Culture and Tourism, China \\
$^{3}$Tencent AI Lab, Bellevue, WA \\
\texttt{wangante@stu.xmu.edu.cn, lfsong@global.tencent.com} \\}

\begin{document}
\maketitle
\begin{abstract}
This work studies improving large language model (LLM) generations at inference time by mitigating fact-conflicting hallucinations.
Particularly, we propose a self-endorsement framework that leverages the fine-grained fact-level comparisons across multiple sampled responses.
Compared with prior ensemble methods (e.g., self-consistency \cite{wang2022self,chen2023universal}) that perform response-level selection, our approach can better alleviate hallucinations, especially for longform generation tasks.
Our approach can broadly benefit smaller and open-source LLMs as it mainly conducts simple content-based comparisons.
Experiments on Biographies show that our method can effectively improve the factuality of generations with simple and intuitive prompts across different scales of LLMs.
Besides, comprehensive analyses on TriviaQA and GSM8K demonstrate the potential of self-endorsement for broader application.
\end{abstract}

\section{Introduction}

Recent Large Language Models (LLMs) such as LLaMA \cite{touvron2023llama} and Mistral \cite{jiang2023mistral} take billions of parameters and are trained on huge corpora of text documents with billions of tokens.
As a result, they have demonstrated remarkable capabilities across various tasks such as longform generation, closed book QA and math reasoning.
However, LLMs can still fail frequently on these knowledge-intensive and reasoning tasks where obviously incorrect facts or reasoning steps are generated.
%
To address this issue, previous work has explored multiple orthogonal directions, such as introducing external knowledge and tool \cite{mallen2023not,peng2023check,wang2023math}, continual supervised finetuning \cite{wu2023eva,tian2023fine} and inference-time improvement \cite{dhuliawala2023chainofverification,chen2023universal} to reduce hallucination and improve reasoning capability.
Among these research directions, inference-time improvement has recently gained popularity.
The motivation behind may stem from various reasons: it can be used on black-box LLMs (e.g., no requirement on accessing the model weighs); it can work together with supervised finetuning by producing high-quality training data (a.k.a., self-distillation \cite{huang2022large}).

\begin{figure*}[t]
    \centering
    \includegraphics[width=0.9\textwidth]{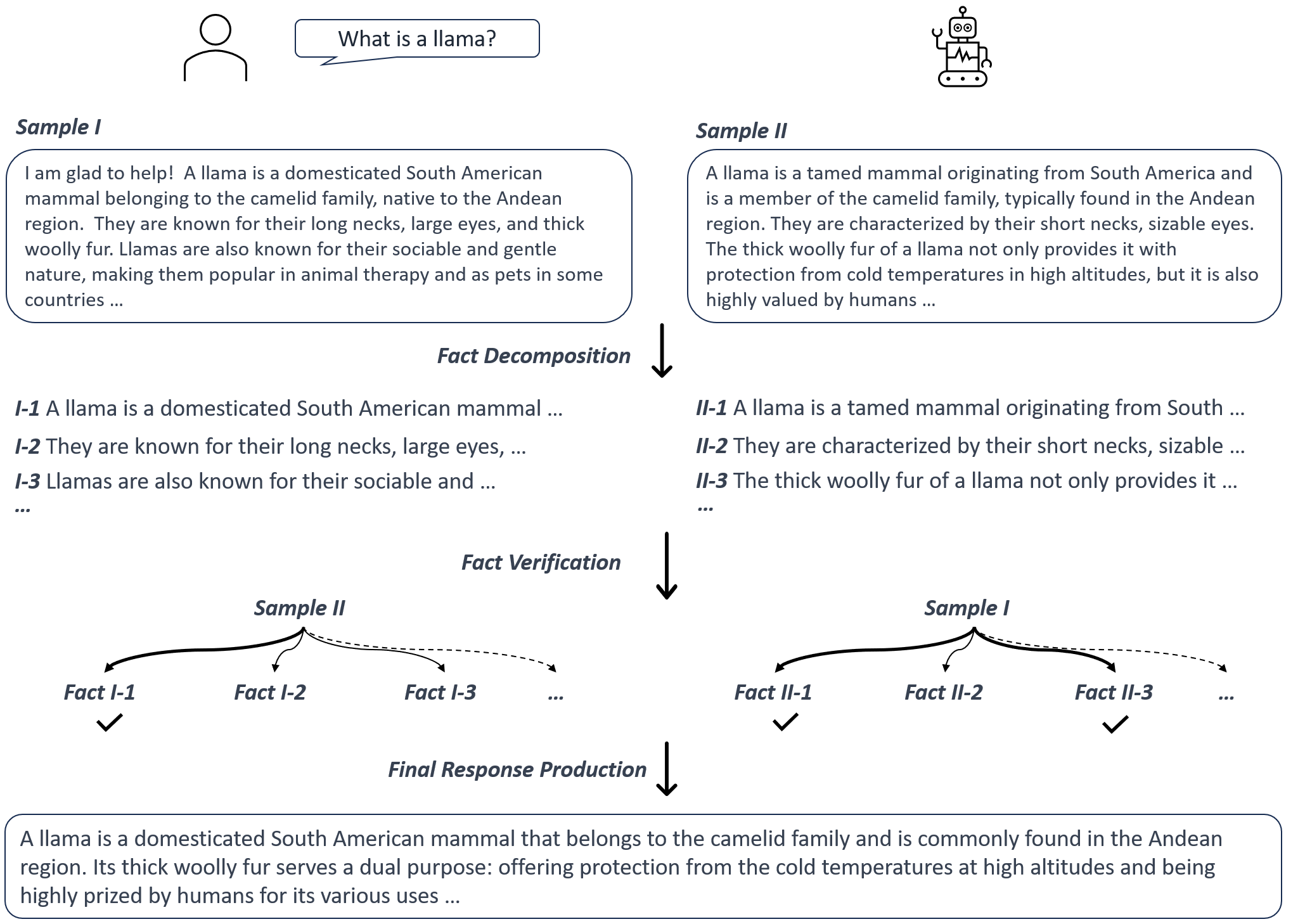}
    \caption{The example framework of self-endorsement, where only two sampled candidates are leveraged.}
    \label{fig:self_endorsement}
\end{figure*}

Many prior approaches of inference-time improvement can be grouped into two main directions.
The \emph{ensemble} methods like self-consistency \cite{wang2022self} and universal self-consistency \cite{chen2023universal} build upon traditional ensemble learning by picking the optimal prediction from multiple candidates sampled from the target LLM.
Conversely, in the other direction, \emph{self-refinement} methods such as chain-of-verification \cite{dhuliawala2023chainofverification} and self-reflection \cite{madaan2023self,shinn2023reflexion} leverage the target LLM to refine its own predictions from varied perspectives.
Comparatively, the ensemble methods can eliminate occasional hallucinations by looking into multiple peering samples.
But, they may fail on longform generation tasks because the sampled candidates disagree with each other on too many places, making it difficult to pick the best prediction.
More importantly, they cannot combine the merits from the peering samples.
On the other hand, the self-refinement methods perform fine-grained refinement.
But they rely on the assumption that the target LLM is strong enough to provide helpful critique for refinement, and thus most experiments on them are conducted on state-of-the-art close-source LLMs (e.g., GPT4 \cite{achiam2023gpt}).

In this work, we follow the line of inference-time improvement to study how and when fine-grained cross-response validation (endorsement) can reduce hallucination and improve reasoning quality.
Particularly, we propose a framework to improve LLM predictions by leveraging fine-grained cross-response endorsements.
As shown in Figure \ref{fig:self_endorsement}, we begin by generating multiple samples from the target LLM.
Next, we extract facts from each sample and prompt the LLM to verify the endorsement of each fact by cross-referencing with the other samples.
An endorsement score is then assigned to each fact based on its level of approval.
Finally, to produce the final response, we either select the sample with the most reliable facts or regenerate a new one by incorporating the facts with high endorsement scores as supplementary inputs to the LLM.
Without complex instructions, the LLM is only required to conduct two tasks: 1) check whether a fact is consistent with the knowledge in another response at a time; 2) generate a new response given additional high-quality facts as inputs.
Both tasks are fairly simple, thus we believe (and our experiments show that) our method can be broadly helpful to various open-source LLMs of different capacities.

We mainly conduct experiments on Biographies \cite{min2023factscore}, a popular benchmark to examine the level of fact-conflicting hallucinations in model predictions for longform generations.
Results on LLaMA2 family models show that our method greatly outperforms baselines by a large margin.
Details analyses suggest that our method can better select reliable fine-grained facts across various model sizes.
We also extensively study on TriviaQA \cite{joshi2017triviaqa} and GSM8K \cite{cobbe2021training}, validating the promise of self-endorsement for more pervasive use.

\section{Baselines}

We take (universal) self-consistency \cite{wang2022self,chen2023universal} and chain-of-verification \cite{dhuliawala2023chainofverification} as the baseline for comparison.
They are two popular methods of inference-time improvement based on ensemble learning and self-refinement, respectively.

\begin{figure}[t]
    \centering
    \begin{subfigure}{.48\textwidth}
        \centering
        \includegraphics[width=1.\linewidth]{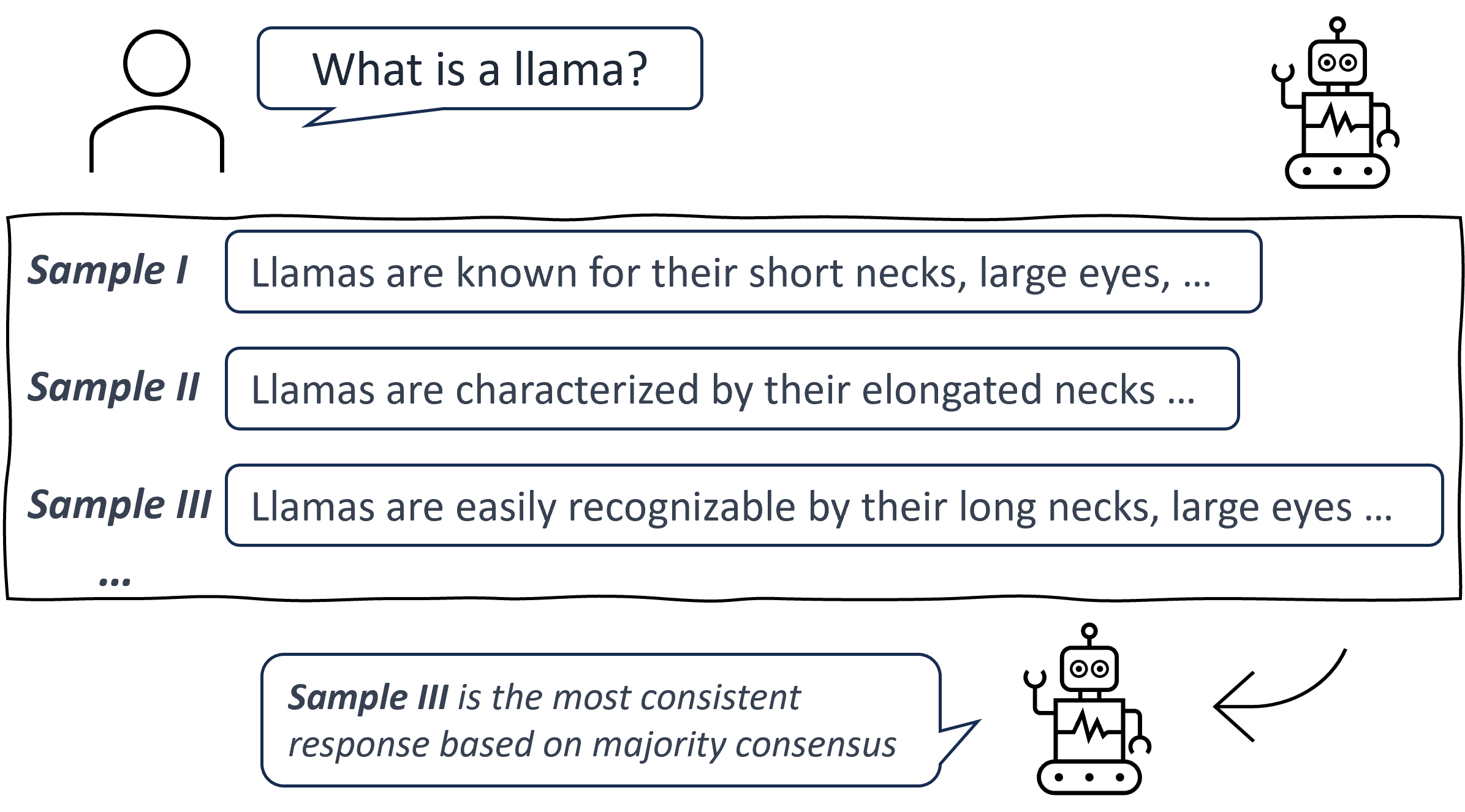}
        \caption{Universal Self-Consistency}
        \vspace{0.5em}
        \label{fig:self-consistency}
    \end{subfigure}
    \begin{subfigure}{.48\textwidth}
        \centering
        \includegraphics[width=0.9\linewidth]{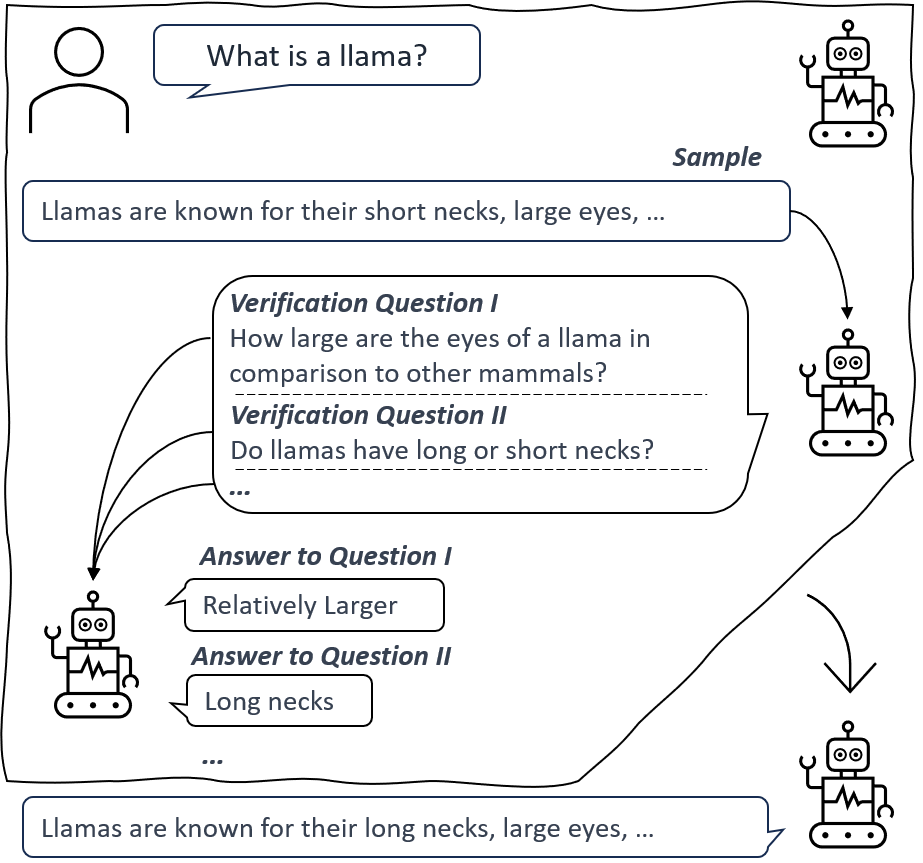}
        \caption{Chain-of-Verification}
        \label{fig:cove}
    \end{subfigure}
    \caption{Two main baselines in this work.}
    \label{fig:pipeline}
    \vspace{-1em}
\end{figure}

\subsection{(Universal) Self-Consistency}
\label{sec:usc}

Self-consistency (SC) is a majority-voting-based ensemble approach designed for reasoning tasks.
Specifically, it first samples multiple reasoning paths and their corresponding answers from the LLM,
e.g., ($r_i, a_i$), where $r_i \Rightarrow a_i$.
It then selects the most consistent answer
via taking a majority vote over $a_i$, 
i.e., $max_{\mathrm{a}} \sum_{i} \mathbbm{1}(a_i = \mathrm{a})$.
With chain-of-thought prompting (CoT), it has demonstrated remarkable performance gains on complex reasoning tasks.
However, self-consistency can only be applied to tasks where the final answer can be aggregated via exact match (e.g., question answering and math word problems).

To support broader applications, universal self-consistency (USC) extends self-consistency by taking the LLM itself (instead of majority voting) to select the final response from the samples it generated.
Particularly as shown in Figure \ref{fig:self-consistency},
the LLM is first asked to sample multiple candidates, it then consumes all these candidates to pick one as the final response.
To achieve precise final answer selection, USC may require that the LLM possesses robust critical analysis capabilities.

\subsection{Chain-of-Verification}
\label{sec:CoVe}
Different from ensemble-based SC\,/\,USC, chain-of-verification (CoVe) refines factual errors in one response and then regenerates a new one by the LLM itself.
As shown in Figure \ref{fig:cove}, the LLM is asked to first (I) draft an initial sample; then (II) plan verification questions to fact-check its draft; (III) answer those questions independently; and (IV) generate its final verified response. 

The core motivation of CoVe is that LLMs tend to provide more accurate facts to simple questions (e.g., the verification questions) than complex questions (e.g., the original question).
Hence it can improve the factuality of the overall response.

\section{Self-Endorsement}

As shown in Figure \ref{fig:self_endorsement}, our self-endorsement framework interacts with an LLM by taking the following steps given a user query $\mathcal{X}$:

\noindent
(1) \emph{Candidate Sampling}: It asks the LLM to sample $N$ candidate responses $Y_1, Y_2, ..., Y_N$.

\noindent
(2) \emph{Fact Decomposition}: It breaks down each candidate $Y_i$ into facts $f^i_1, f^i_2, ..., f^i_{N_{Y_i}}$, where $N_{Y_i}$ is the number of facts in $Y_i$.

\noindent
(3) \emph{Fact Verification}: It verifies each fact $f^i_j$ via calculating its endorsement scores against other candidates $\{Y_k~|~k\neq i\}$. We also explore context pruning, which eliminates unrelated content in candidates for verification.

\noindent
(4) \emph{Final Response Production}: Produce a final response via selection or regeneration. Specifically, we either select the response with facts having the highest endorsement scores as the final response or ask the LLM to regenerate a new one $Y$ given the set of selected facts 
$\mathcal{Z}$
from different candidates.


\subsection{Candidate Sampling}
\label{sec:sample}
We follow the common practice of sampling $N$ responses via nucleus sampling. Each sampling process is denoted as $Y_i \sim \texttt{LLM}(\mathcal{X})$.

\subsection{Fact Decomposition}
Following exiting work \cite{gao2022attributed,liu2023evaluating}, we consider a fact as a statement about some factual knowledge.
There are many ways to conduct fact decomposition.
We first adopt a naive method used by some previous work \cite{liu2023evaluating,manakul2023selfcheckgpt}, which takes each sentence in a response as a fact.
However, it fails to consider the situations that some sentences can contain multiple independent facts \cite{liu2023evaluating} or do not contain any fact.
Therefore, we also study prompting the LLM itself to extract facts from its responses.
This process is denoted as
$f^i_1, f^i_2, ..., f^i_{N_{Y_i}} = \texttt{LLM}(Y_i, P_{D})$, where $P_{D}$ is the corresponding LLM instruction shown below:

\vspace{0.3em}
\noindent\fbox{%
    \parbox{0.97\linewidth}{%
        \emph{List all non-repeated facts from the text below in numerical order. Each fact should be a self-contained sentence: $Y_i$}
    }%
}
\vspace{0.1em}

We observe that the LLM-prompting method can effectively eliminate statements without factual knowledge and break down complex sentences into multiple pieces of facts.

\subsection{Fact Verification by Self-Endorsement}
\label{sec:verify}

We verify each fact via its endorsement score: the degree of the fact being consistent with the content in other sampled responses.
There are multiple ways to compare two pieces of text, such as querying the LLM or calling a sentence encoder (e.g. SimCSE \cite{gao2021simcse}).
For simplicity and to minimize the effect of extra supervision, we choose to query the LLM via prompting.

Formally, for a fact $f^i_j$ from response $Y_i$, we feed $f^i_j$ and another response $Y_k$ ($k \neq i$) to the LLM with prompt $P_{V}$ to determine whether $Y_k$ endorses $f^i_j$.
Then, we define the endorsement score of $f^i_j$ as:
\begin{equation*}
    g(f^i_j) = \frac{1}{N-1} \sum_{k\neq i} \texttt{LLM}(f^i_j, Y_k, P_{V}).
\end{equation*}
The prompt $P_{V}$ is simply defined as:

\vspace{0.3em}
\noindent\fbox{%
    \parbox{0.97\linewidth}{%
        \emph{Take the following as truth: $Y_k$}\\
\emph{Then the following statement: ``$f^i_j$'' is true, false, or inconclusive?}
    }%
}
\vspace{0.1em}

In many situations, especially for longform generation, most facts in $Y_k$ can be irrelevant to $f^i_j$.
Therefore, we propose to further prune the unnecessary context and only keep the most related parts to speed up inference.
Particularly, we select top-$K$ similar facts to $f^i_j$ from each $Y_k$ using the BM25 algorithm.
Then, we concatenate the $K$ selected facts (denoted as $Y'_k$) to verify $f^i_j$.

Generally, 
the endorsement score reflects the level of confidence from the LLM to a piece of fact.
Therefore, facts with higher endorsement scores have higher chances to be faithful.

\subsection{Selection\,/\,Regeneration for Final Response Production}
\label{sec:final}

\paragraph{Selection}
After the above steps, a simple option is to select one from the sampled candidates as the final response $Y$.
For each candidate $Y_i$, we average the endorsement scores of its facts (i.e., $\texttt{Avg}(g(f^i_1),...)$) and select the one with the highest average score as the final response.
However, this does not fully exploit the potential of our framework due to the following reasons:
(1) There can still be factual errors in the selected response.
(2) Helpful and complementary facts in other responses are not efficiently leveraged.

\paragraph{Regeneration}
We propose another option that prompts the LLM to regenerate the final response $Y$ with selected facts ($\mathcal{Z}$)
from all samples: 
$Y \sim \texttt{LLM}(\mathcal{X}, \mathcal{Z}, P_{G})$, where prompt $P_{G}$ is defined as:

\vspace{0.3em}
\noindent\fbox{%
    \parbox{0.97\linewidth}{%
        \emph{Knowledge from other sources: $\mathcal{Z}$}\\
\emph{Given the materials above, answer the question: $\mathcal{X}$}
    }%
}
\vspace{0.1em}

To select useful facts, we first discard the facts whose endorsement scores do not exceed a threshold $\alpha$ (i.e., $g(f^i_j) \leq \alpha$).
Though this can effectively prune low-quality facts, there can still be facts of redundant content.
We then adopt a K-means algorithm that takes bag-of-words features as the representation for each fact and groups the facts into $\mathcal{C}$ clusters.
Lastly, we select the fact closest to the centroid for each cluster to form the selected fact set $\mathcal{Z}$ that contains $|\mathcal{C}|$ facts.


\section{Experiments}

\subsection{Setup}

\paragraph{Datasets}
We mainly conduct experiments on Biographies \cite{min2023factscore}. 
It contains 183 person entities used to prompt LLMs about their biographies with the query \emph{``Tell me a bio of <entity>''}.
As the responses of LLMs can be long and contain a wealth of factual knowledge, it has been a popular benchmark for evaluating factuality in longform text generation \cite{dhuliawala2023chainofverification,tian2023fine}.
In addition to that, we also test self-endorsement on a popular QA benchmark TriviaQA \cite{joshi2017triviaqa} and a math dataset GSM8K \cite{cobbe2021training}.
More details about both datasets are introduced later in this section.

\paragraph{Evaluation}
For Biographies, we follow \citet{min2023factscore} to evaluate the accuracy of decomposed facts (\emph{Fact Acc.}) using their released inst-LLaMA-7B model
together with the Wikipedia dump from 2023/04/01 as judge.
Particularly, the correctness of each fact is evaluated by inst-LLaMA-7B that takes the top 5 passages retrieved from
the wiki page of the topic entity as extra evidence.
Though inst-LLaMA-7B is much smaller than the start-of-the-art LLMs such as ChatGPT, \citet{min2023factscore} has shown that inst-LLaMA-7B can always give consistent judging decisions with ChatGPT.
In addition to \emph{Fact Acc.}, we also report the number of facts (\emph{\#Fact}),
because good responses should contain a decent number of facts of high accuracy.

For TriviaQA, we follow standard practice to also report answer recall (\emph{Ans. Rec.})
in addition to fact accuracy and the number of facts.
Answer recall measures if the target answer is contained in the generated response.
For GSM8K, we report the quality of the intermediate reasoning steps using GPT4 as judge (\emph{GPT4 (Y)} and \emph{GPT4 (N)}) in addition to the accuracy of the final answer (\emph{Acc.}).
More details on the quality of the intermediate steps are introduced in the corresponding section.

\paragraph{Settings and Hyperparameters}
We conduct experiments based on LLaMA2-7B-Chat and LLaMA2-70B-Chat \cite{touvron2023llama} for Biographies and TriviaQA.
Mixtral-8x7B-Inst \cite{jiang2023mistral} is adopted for GSM8K due to its stronger math capabilities.

For our approach, we use nucleus sampling with a temperature of 1.0 when generating responses and use greedy decoding otherwise.
We prompt the target LLM to extract facts for Biographies and TriviaQA and directly take each sentence in a response as a fact for GSM8K.
We empirically set candidate number $N$ (\S \ref{sec:sample}), the number of kept context facts $K$ (\S \ref{sec:verify}), and fact-filtering threshold $\alpha$ (\S \ref{sec:final}) as 10\,/\,10, 3\,/\,3, 1.0\,/\,0.8 for LLaMA2-7B-Chat\,/\,LLaMA2-70B-Chat.
The K-means cluster number $\mathcal{C}$ is dynamically decided by the average number of facts across the $N$ candidate responses.  
We also conduct careful analyses on the effects of these hyperparameters.

\paragraph{Baselines}
One obvious baseline is simply calling LLM to sample a response.
We report the average numbers from $N$ sampled responses to alleviate the randomness of the sampling process.
In addition, we take the following baselines for a better understanding of our approach:
\begin{itemize}[leftmargin=*]
    \item \emph{Refine}: Considering the power of LLMs, an LLM might be able to correct its own errors given a second chance. This baseline is set to quantify the gain from this effect.
    \item \emph{(Universal) Self-Consistency (SC\,/\,USC)}: They are implemented as mentioned in \S \ref{sec:usc}.
    \item \emph{Chain-of-Verification (CoVe)}: Its implementation follows the description in \S \ref{sec:CoVe}.
\end{itemize}


\subsection{Results and Analyses}

\begin{table}[]
\small
    \centering
    \begin{tabular}{lcc}
    \toprule
    Model & Fact Acc. & \#Fact \\
    \hline
    LLaMA2-7B-Chat & 53.2 & 16.8 \\
    ~~+refine & 52.6 & 15.7 \\
    ~~+USC & 53.5 & 15.9 \\
    ~~+CoVe & 54.8 & 9.8 \\
    \hdashline
    {\color{gray} \emph{self-endorsement}} & & \\
    ~~+select & \phantom{**}58.2** & 15.9 \\
    ~~+select w/ pruning & \phantom{**}59.6** & 15.2 \\
    ~~+regenerate & \phantom{**}\textbf{67.7}** & 14.9 \\
    ~~+regenerate w/ pruning & \phantom{**}65.7** & 14.6 \\
    \hline
    \hline
    LLaMA2-70B-Chat & 63.1 & 20.0 \\
    ~~+refine & \phantom{*}64.9* & 20.2 \\
    ~~+USC & 61.6 & 20.4 \\
    ~~+CoVe & 64.0 & 16.5 \\
    \hdashline
    {\color{gray} \emph{self-endorsement}} & & \\
    ~~+select & \phantom{**}66.5** & 19.4 \\
    ~~+select w/ pruning & \phantom{**}67.7** & 18.8 \\
    ~~+regenerate & \phantom{**}\textbf{73.1}** & 18.3 \\
    ~~+regenerate w/ pruning & \phantom{**}73.0** & 17.9 \\
    \bottomrule
    \end{tabular}
    \caption{Test results on Biographies. We also report significant test results using bootstrap resampling. *, ** denote significantly better results over the base LLM (the first line in each group) with significance level $p < 0.05$ and $p < 0.01$, respectively.}
    \label{tab:main_result}
\end{table}

\paragraph{\emph{Self-endorsement Helps Improving Factuality}}
As shown in Table \ref{tab:main_result}, none of the baselines (\emph{+refine}, \emph{+USC} and \emph{+CoVe}) can significantly improve over the 7B and 70B LLaMA2-Chat model regarding \emph{Fact Acc.}
In contrast, self-endorsement gives significant improvements over baselines no matter whether the final response is selected or regenerated and whether context pruning is used or not.

Among those baselines, only \emph{CoVe} can slightly improve \emph{Fact Acc.}, but it obviously decreases the \emph{\#Fact}, which is also observed in \citet{dhuliawala2023chainofverification}.
\emph{Refine} only benefits LLaMA2-70B-Chat, 
while the gain is still much inferior to our self-endorsement approaches based on self-selected high-quality facts.
The results of \emph{Refine} also indicate that naive self-refinement demands strong capabilities of the LLM.


For our methods, because regeneration can include reliable facts from all candidates and discard incorrect facts, thus it consistently produces better responses than selection.
Using context pruning or not gives a minor performance change regarding \emph{Fact Acc}. We will provide more analyses in later experiments.

\begin{figure}[t]
    \begin{subfigure}{.235\textwidth}
        \centering
        \includegraphics[width=1.05\linewidth]{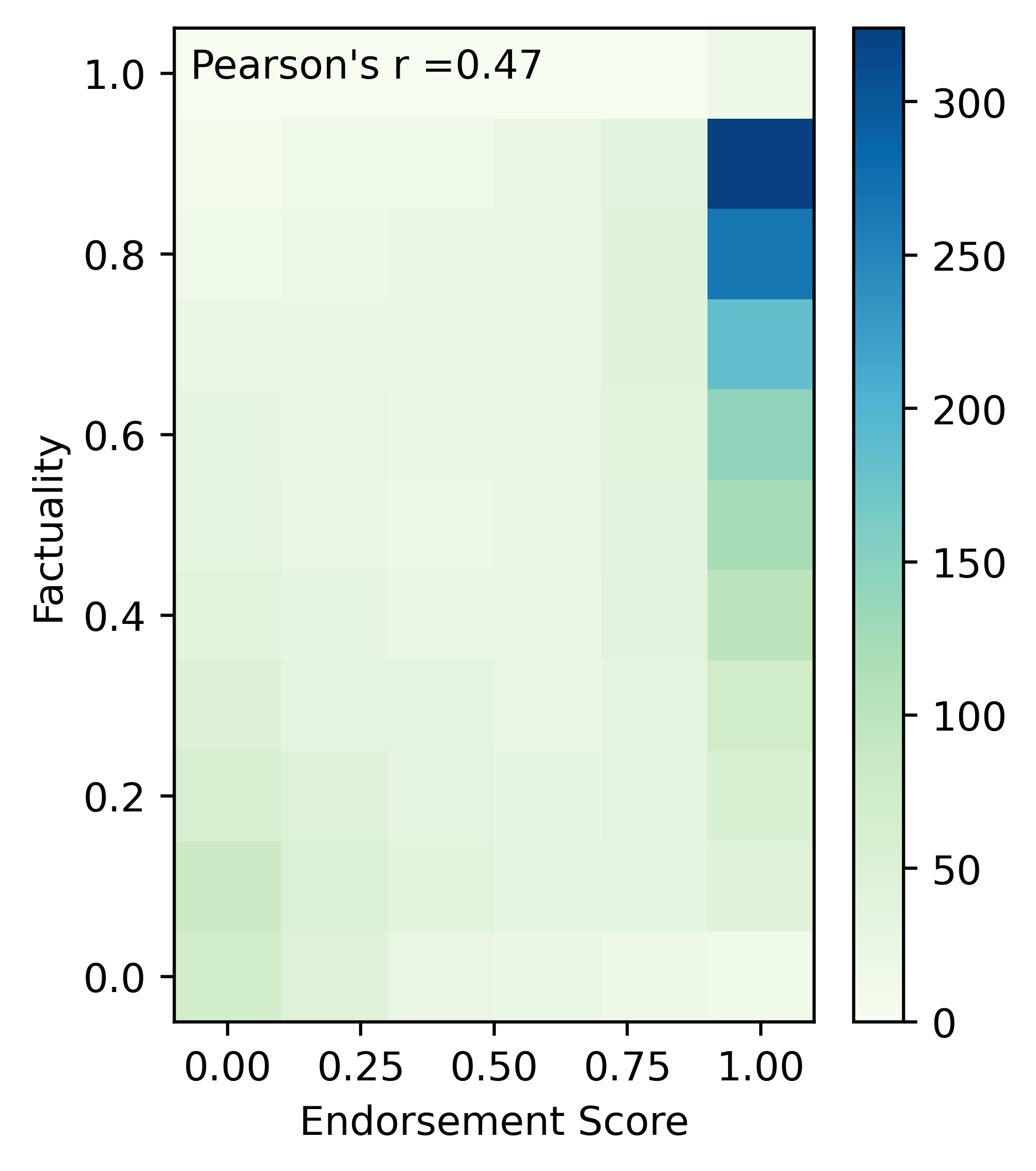}
        \caption{LLaMA2-7B-Chat}
        \label{fig:7b_endorse}
    \end{subfigure}
    \begin{subfigure}{.235\textwidth}
        \centering
        \includegraphics[width=1.05\linewidth]{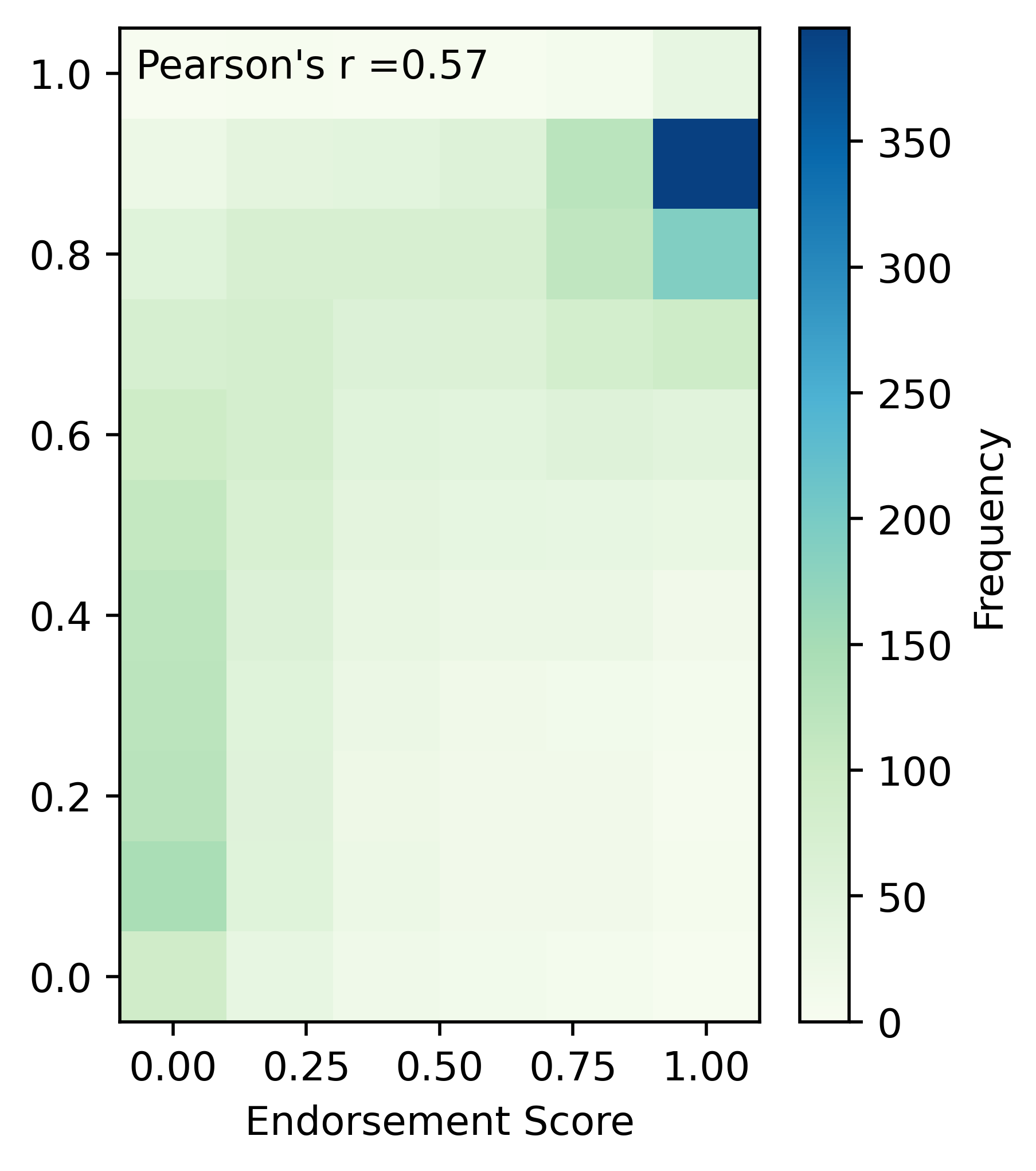}
        \caption{LLaMA2-70B-Chat}
        \label{fig:70b_endorse}
    \end{subfigure}
    \caption{Statistical correlation between endorsement scores and factuality scores.} 
    \label{fig:endorse_factual}
\end{figure}

\begin{figure*}[t!]
    \centering
    \begin{subfigure}{.32\textwidth}
        \centering
        \begin{tikzpicture}[scale=0.57]
        \begin{axis}[
            axis y line*=left,
            xlabel=$\alpha$,
            ylabel=Fact Acc. (\%) \ref{pgfplots:plot1},
            ymin=55, ymax=70,
            ylabel near ticks,
            ylabel style={anchor=near ticklabel, align=center},
        ]
        \addplot[blue,mark=*] coordinates {
            (0, 59.2)
            (0.2, 62.7)
            (0.4, 62.8)
            (0.6, 63.3)
            (0.8, 65.7)
            (1.0, 67.7)
        };
        \label{pgfplots:plot1}
        \end{axis}
        \begin{axis}[
            axis y line*=right,
            ymin=10, ymax=20,
            ylabel near ticks,
            ylabel style={anchor=near ticklabel, align=center},
        ]
        \addplot[gray,dashed,mark=square*] coordinates {
            (0, 14.8)
            (0.2, 14.8)
            (0.4, 14.9)
            (0.6, 15.1)
            (0.8, 14.4)
            (1.0, 14.9)
        };
        \label{pgfplots:plot2}
        \end{axis}
        \end{tikzpicture}
        \caption{}
        \label{fig:7b_alpha}
    \end{subfigure}
    \begin{subfigure}{.32\textwidth}
        \centering
        \begin{tikzpicture}[scale=0.57]
        \begin{axis}[
            axis y line*=left,
            xlabel=$N$,
            ymin=60, ymax=70,
            ylabel near ticks,
            ylabel style={anchor=near ticklabel, align=center},
        ]
        \addplot[blue,mark=*] coordinates {
            (2, 62.3)
            (4, 63.8)
            (6, 65.9)
            (8, 66.0)
            (10, 67.7)
        };
        \end{axis}
        \begin{axis}[
            axis y line*=right,
            ymin=10, ymax=20,
            ylabel near ticks,
            ylabel style={anchor=near ticklabel, align=center},
        ]
        \addplot[gray,dashed,mark=square*] coordinates {
            (2, 14.1)
            (4, 14.3)
            (6, 14.4)
            (8, 15.0)
            (10, 14.9)
        };
        \end{axis}
        \end{tikzpicture}
        \caption{}
        \label{fig:7b_candidate_num}
    \end{subfigure}
    \begin{subfigure}{.32\textwidth}
        \centering
        \begin{tikzpicture}[scale=0.57]
        \begin{axis}[
            axis y line*=left,
            xlabel=$M$,
            ymin=62, ymax=70,
            ylabel near ticks,
            ylabel style={anchor=near ticklabel, align=center},
        ]
        \addplot[blue,mark=*] coordinates {
            (2, 64.2)
            (4, 65.1)
            (6, 65.6)
            (8, 66.0)
            (10, 67.7)
        };
        \end{axis}
        \begin{axis}[
            axis y line*=right,
            ylabel=\#Fact \ref{pgfplots:plot2},
            ymin=10, ymax=20,
            ylabel near ticks,
            ylabel style={anchor=near ticklabel, align=center},
        ]
        \addplot[gray,dashed,mark=square*] coordinates {
            (2, 14.2)
            (4, 14.3)
            (6, 14.4)
            (8, 14.6)
            (10, 14.9)
        };
        \end{axis}
        \end{tikzpicture}
        \caption{}
        \label{fig:7b_fact_num}
    \end{subfigure}
    \begin{subfigure}{.32\textwidth}
        \centering
        \begin{tikzpicture}[scale=0.57]
        \begin{axis}[
            axis y line*=left,
            xlabel=$\alpha$,
            ylabel=Fact Acc. (\%) \ref{pgfplots:plot1},
            ymin=65, ymax=75,
            ylabel near ticks,
            ylabel style={anchor=near ticklabel, align=center},
        ]
        \addplot[blue,mark=*] coordinates {
            (0, 66.1)
            (0.2, 72.3)
            (0.4, 73.5)
            (0.6, 73.7)
            (0.8, 73.1)
            (1.0, 71.7)
        };
        \label{pgfplots:plot1}
        \end{axis}
        \begin{axis}[
            axis y line*=right,
            ymin=14, ymax=22,
            ylabel near ticks,
            ylabel style={anchor=near ticklabel, align=center},
        ]
        \addplot[gray,dashed,mark=square*] coordinates {
            (0, 18.3)
            (0.2, 18.3)
            (0.4, 18.3)
            (0.6, 18.1)
            (0.8, 18.0)
            (1.0, 18.2)
        };
        \label{pgfplots:plot2}
        \end{axis}
        \end{tikzpicture}
        \caption{}
        \label{fig:70b_alpha}
    \end{subfigure}
    \begin{subfigure}{.32\textwidth}
        \centering
        \begin{tikzpicture}[scale=0.57]
        \begin{axis}[
            axis y line*=left,
            xlabel=$N$,
            ymin=68, ymax=76,
            ylabel near ticks,
            ylabel style={anchor=near ticklabel, align=center},
        ]
        \addplot[blue,mark=*] coordinates {
            (2, 70.4)
            (4, 72.0)
            (6, 72.8)
            (8, 73.9)
            (10, 73.7)
        };
        \end{axis}
        \begin{axis}[
            axis y line*=right,
            ymin=14, ymax=22,
            ylabel near ticks,
            ylabel style={anchor=near ticklabel, align=center},
        ]
        \addplot[gray,dashed,mark=square*] coordinates {
            (2, 18.0)
            (4, 17.9)
            (6, 17.8)
            (8, 18.0)
            (10, 18.1)
        };
        \end{axis}
        \end{tikzpicture}
        \caption{}
        \label{fig:70b_candidate_num}
    \end{subfigure}
    \begin{subfigure}{.32\textwidth}
        \centering
        \begin{tikzpicture}[scale=0.57]
        \begin{axis}[
            axis y line*=left,
            xlabel=$M$,
            ymin=68, ymax=76,
            ylabel near ticks,
            ylabel style={anchor=near ticklabel, align=center},
        ]
        \addplot[blue,mark=*] coordinates {
            (2, 71.3)
            (4, 72.8)
            (6, 71.8)
            (8, 73.1)
            (10, 73.7)
        };
        \end{axis}
        \begin{axis}[
            axis y line*=right,
            ylabel=\#Fact \ref{pgfplots:plot2},
            ymin=14, ymax=22,
            ylabel near ticks,
            ylabel style={anchor=near ticklabel, align=center},
        ]
        \addplot[gray,dashed,mark=square*] coordinates {
            (2, 17.7)
            (4, 17.8)
            (6, 18.3)
            (8, 17.9)
            (10, 18.1)
        };
        \end{axis}
        \end{tikzpicture}
        \caption{}
        \label{fig:70b_fact_num}
    \end{subfigure}
    \caption{Hyperparameter analyses on LLaMA-7B-Chat (up) and LLaMA-70B-Chat (down). We present different choices of $\alpha$, $N$ and $M$ and their effects on \emph{Fact Acc.} and \emph{\#Fact}.}
    \label{fig:param_select}
\end{figure*}
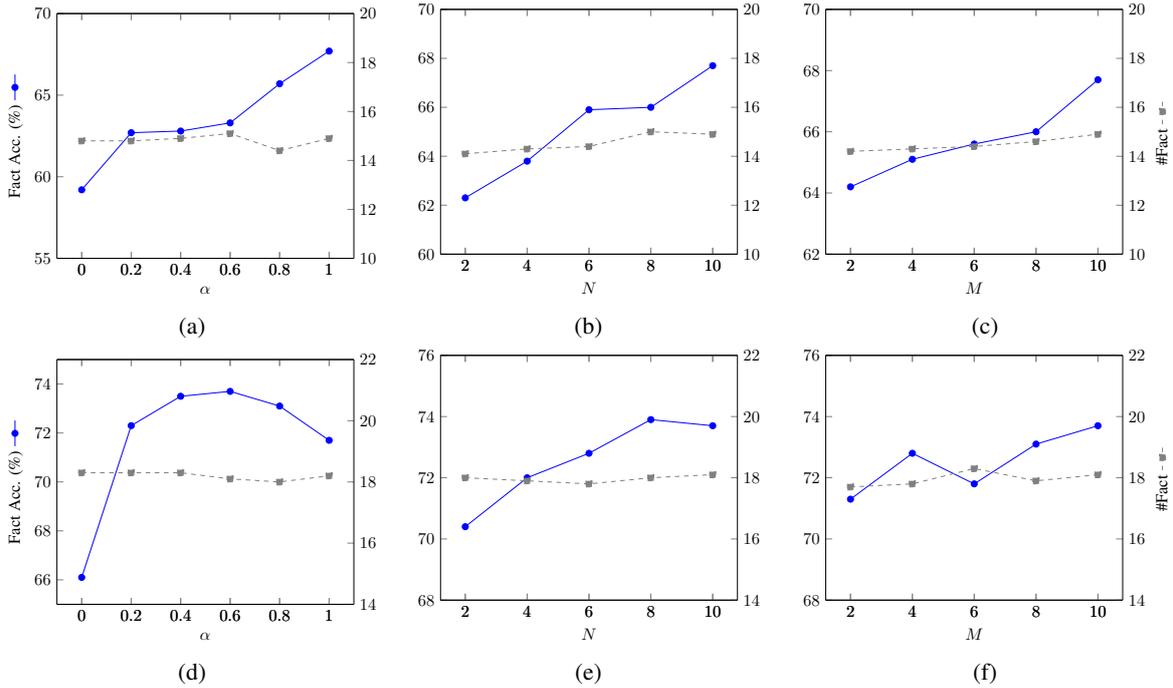

\paragraph{\emph{Endorsement Score Correlates with Factuality}}
Since endorsement scores play a crucial role in the success of our approaches, we further investigate how endorsement scores are correlated with the actual factuality.
To this end, we use inst-LLaMA-7B with Wikipedia dump to calculate the factuality score for each piece of fact.
Figure \ref{fig:endorse_factual} presents the correlation between endorse scores and factuality scores.
Results on both models show clear positive relationships between endorsement scores and factuality.
LLaMA2-70B-Chat gives a stronger correlation because of its stronger ability over LLaMA2-7B-Chat.
Especially, LLaMA2-7B-Chat tends to give higher endorsement scores to certain incorrect facts erroneously.

\paragraph{\emph{How the Quality of Selected Facts Affect Final Responses?}}
Since threshold $\alpha$ decides the quality of selected facts for regeneration, here we try several values of $\alpha$ and visualize the corresponding final-response quality in
Figure \ref{fig:7b_alpha} and \ref{fig:70b_alpha}.
We observe that ranging $\alpha$ from 0 to 1 keeps benefiting LLaMA2-7B-Chat but the performance on LLaMA2-70B-Chat is increased first and then decreased.
After a closer look, we find that a high $\alpha$ may limit the quantity and diversity of selected facts, which may hurt the regeneration quality.
For example, when $\alpha=1$, only an average number of 11.3 facts are selected under LLaMA2-70B-Chat, while the number is 16.7 for LLaMA2-7B-Chat.
Besides, we observe a decent performance increase with $\alpha \ge 0.2$, showing the effectiveness of our approach on alleviating the side-effect of low-quality facts by removing them.

\paragraph{\emph{How Candidate Number Affects Final Responses?}}
Intuitively, increasing the candidate number $N$ can help to provide more high-quality facts and each fact can also be better verified with more samples.
As shown in Figure \ref{fig:7b_candidate_num} and \ref{fig:70b_candidate_num}, the performances of both 7B and 70B models generally get improved when increasing $N$, and the number of facts in regenerated responses remains stable.
For LLaMA2-7B-Chat, more improvements can be expected when $N$ is further increased.
However, this will also bring more computational costs that can be impractical.
In contrast, LLaMA2-70B-Chat is less sensitive, showing that a small $N$ is enough for stronger LLMs.
Encouragingly, we also observe that our models can significantly outperform baselines with limited samples (70.4 vs. 63.1 when $N=2$ on LLaMA2-70B-Chat).
This suggests the robustness of our method in some extreme cases.

\paragraph{\emph{Effect on Selecting Facts from Fewer Candidates for Regeneration}}
We further analyze the effect of selecting facts from fewer number (denoted as $M$ and $M < N$) of candidates.
Note that these facts from the $M$ candidates can still take all $N$ candidates to calculate their endorsement scores.
Results are shown in Figure \ref{fig:7b_fact_num} and \ref{fig:70b_fact_num}.
We again observe positive effects when increasing $M$, because the final responses can directly consult more provided input facts.
Besides, by comparing the results in Figure \ref{fig:7b_candidate_num} and \ref{fig:7b_fact_num} (also Figure \ref{fig:70b_candidate_num} vs \ref{fig:70b_fact_num}), we find that the latter performs better when the candidate number is small (e.g., 71.3 vs. 70.4 when both $N=2$ and $M=2$ on LLaMA2-70B-Chat).
This indicates that a fact can be better verified when more candidates are available for calculating endorsement scores.

\begin{table}[]
\small
    \centering
    \begin{tabular}{ccc}
    \toprule
    $K$ & Fact Acc. & \#Fact \\
    \hline
    1 & 62.5 & 15.1 \\
    3 & 65.7 & 14.6 \\
    5 & 66.8 & 14.7 \\
    \hdashline
    ALL & \textbf{67.7} & 14.9 \\
    \hline\hline
    1 & 72.4 & 18.1 \\
    3 & 73.0 & 17.9 \\
    5 & \textbf{73.2} & 18.2 \\
    \hdashline
    ALL & 73.1 & 18.3 \\
    \bottomrule
    \end{tabular}
    \caption{Performances on LLaMA2-7B-Chat (up) and LLaMA2-70B-Chat (down) when using $K$ facts from other responses to calculate the endorsement score for target facts.}
    \label{tab:context_pruning}
\end{table}

\paragraph{\emph{How Context Pruning Affects Final Responses?}}
Context pruning aims to eliminate unnecessary context when calculating the endorsement score for each fact,
while it may hurt the accuracy of fact selection and overall performance when too much context is pruned.
As shown in Table \ref{tab:context_pruning} (up),
LLaMA2-7B-Chat is largely influenced by $K$, and its performance stably improves when $K$ increases.
Conversely, though growing \emph{Fact Acc.} scores are observed as well for LLaMA2-70B-Chat (Table \ref{tab:context_pruning} (down)), the growth rate is mild (e.g., 72.4 $\rightarrow$ 73.2).
This is consistent with the comparison on both candidate number $N$ (Figure \ref{fig:7b_candidate_num} vs \ref{fig:70b_candidate_num}) and candidate number for fact selection $M$ (Figure \ref{fig:7b_fact_num} vs \ref{fig:70b_fact_num}).
For both 7B and 70B models, we observe \emph{Fact Acc.} numbers that are close to when no context pruning is used.
Thus, context pruning is useful overall, especially considering that it can save 50\% of computation cost when $K=5$ according to statistics.
Note that we only use the vanilla BM25 algorithm for selecting related facts.
We leave exploring better sentence matching algorithms in future work.


\begin{table}[h]
\setlength{\tabcolsep}{2pt}
\small
    \centering
    \begin{tabular}{lccc}
    \toprule
    Model & Fact Acc. & Ans. Rec. & \#Fact \\
    \hline
    LLaMA2-7B-Chat & 57.4 & 70.0 & 4.8 \\
    ~~+USC & 57.6 & 69.0 & 4.8 \\
    ~~+CoVe & 53.7 & \textbf{71.2} & 4.3 \\
    \hdashline
    {\color{gray} \emph{self-endorsement}} & & & \\
    ~~+select & \phantom{**}63.4** & 70.2 & 4.4 \\
    ~~+select w/ pruning & \phantom{**}63.8** & 69.5 & 4.3 \\
    ~~+regenerate & \phantom{**}\textbf{65.0}** & 70.7 & 4.7 \\
    ~~+regenerate w/ pruning & \phantom{**}64.0** & 70.8 & 4.4 \\
    \hline
    \hline
    LLaMA2-70B-Chat & 65.1 & 84.1 & 5.0 \\
    ~~+USC & 65.0 & 83.1 & 5.0 \\
    ~~+CoVe & 58.9 & 83.1 & 5.4 \\
    \hdashline
    {\color{gray} \emph{self-endorsement}} & & & \\
    ~~+select & \phantom{**}69.7** & 83.8 & 4.8 \\
    ~~+select w/ pruning & \phantom{**}70.2** & 84.2 & 4.7 \\
    ~~+regenerate & \phantom{**}\textbf{71.7}** & \phantom{*}\textbf{85.3}* & 5.2 \\
    ~~+regenerate w/ pruning & \phantom{**}70.7** & \phantom{*}85.0* & 5.2 \\
    \bottomrule
    \end{tabular}
    \caption{Test results on TriviaQA.}
    \label{tab:triviaqa}
\end{table}

\paragraph{\emph{Evaluation Results on Question Answering}}
To validate our approach to short-text generation, we then conduct experiments on TriviaQA \cite{joshi2017triviaqa}, a popular open-domain question-answering benchmark.
We do not add restrictions (e.g., early stopping or instructing the LLM to only generate the answer) to encourage the LLM to generate explanations and relevant knowledge in addition to the answer.
For evaluation, we report answer recall (\emph{Ans. Rec.}) in addition to \emph{Fact Acc.} and \emph{\#Fact}.
We randomly sample 1000 questions from the original development set of the Wikipedia domain.

Results are shown in Table \ref{tab:triviaqa}.
Our method again effectively improves the \emph{Fact Acc.}, which is consistent with our observations on Biographies.
The improvements regarding \emph{Ans. Rec.} are limited.
It is because LLMs have already provided more accurate exact answers to target questions \cite{dhuliawala2023chainofverification} but tend to ignore other facts in the responses.
Besides, \emph{regeneration} gives fewer improvements over \emph{selection} on this dataset, which can be due to the limited fact numbers in short-text generation thus \emph{selection} is also easier to select a good one from enough candidates.

\begin{table}[]
\setlength{\tabcolsep}{3pt}
\small
    \centering
    \begin{tabular}{lccc}
    \toprule
    Model & Acc. & GPT4 (Y) & GPT4 (N) \\
    \hline
    Mixtral-8$\times$7B-Inst & 68.4 & 9.87 & 3.65 \\
    ~~+USC & \phantom{*}71.6* & 9.86 & \phantom{*}3.90* \\
    ~~+CoVe & 56.0 & -- & -- \\
    ~~+SC & \phantom{**}80.3** & 9.87 & \phantom{**}3.96** \\
    ~~+select & \phantom{**}\textbf{80.8}** & 9.87 & \phantom{**}\textbf{4.08}** \\
    \bottomrule
    \end{tabular}
    \caption{Test results on GSM8K. We do not report \emph{CoVe} results on GPT4 because its answers usually do not contain complete rationales.}
    \label{tab:gsm8k}
\end{table}

\paragraph{\emph{Extensive Experiments on GSM8K}}
In addition to knowledge-intensive tasks, we also briefly explore self-endorsement on reasoning tasks, choosing GSM8K \cite{cobbe2021training}, a popular math benchmark, as the testbed.
Here we focus more on the quality of intermediate reasoning steps in addition to the final-answer accuracy (\emph{Acc.}).
Particularly, we divide the reasoning steps into two groups (\emph{Yes\,/\,No}) based on whether their corresponding predicted answers are correct or not.
We then prompting \texttt{gpt-4-0613} with the instruction\footnote{See Figure \ref{fig:gpt4_prompt} in Appendix.} from the MT-bench \cite{zheng2023judging} to measure the quality of each group (\emph{GPT4 (Y)}\,/\,\emph{GPT4 (N)}).


As shown in Table \ref{tab:gsm8k}, both \emph{USC} and \emph{SC} help improve \emph{Acc.} while \emph{SC} performs significantly better.
This is because \emph{SC}, which conducts majority voting on final answers, is more aligned with \emph{Acc.}
\emph{CoVe} even severely hurt model performance.
This is because the augmented questions occasionally inquire about irrelevant topics, which disturb the main reasoning procedure.

Regarding the intermediate steps, there is a large performance gap between both groups (\emph{Yes\,/\,No}).
Thus, how to further improve the group of incorrect final answers has become critical.
Our method reports a slightly better result than \emph{SC} on \emph{Acc}, and the gap on \emph{GPT4 (N)} is even more (0.12 over 10).
This indicates that our method indeed helps select relatively better rationales even though the final answers are incorrect, validating the effectiveness of our method from another aspect.

\section{Related Work}

\subsection{Inference-time Hallucination Mitigation}
Researchers have explored mitigating LLM hallucinations at both training and inference time.
Compared with training-time mitigation approaches \cite{lee2022factuality,lightman2023let,tian2023fine}, inference-time improvement is gaining popularity because it can be more cost-effective and controllable \cite{zhang2023siren}.

Except for the two baselines \emph{USC} and \emph{CoVe} we introduced previously, \citet{lee2022factuality} proposed \emph{factual-nucleus sampling} that balances diversity and factuality by dynamically adjusting the hyperparameters of sampling when decoding.
\citet{li2023inference} introduced \emph{Inference-Time Intervention (ITI)} that shifts model activations along truth-correlated directions after identifying attention heads with high linear probing accuracy for truthfulness.
\citet{chuang2023dola} found that factual information is encoded in distinct layers, thus they contrasted the generation probabilities from different layers of LLMs.
Among these studies, our approach is most related to \emph{USC} involving checking the consistency across sampled candidates but is conducted at the fact level.

\subsection{Black-box Hallucination Detection}
Detecting hallucinations during inference is usually based on uncertainty estimation.
Current work can be categorized into three types \cite{zhang2023siren}:
logit-based \cite{guo2017calibration}, verbalize-based \cite{xiong2023llms}, and consistency-based \cite{manakul2023selfcheckgpt,mundler2023self}.

This work is most relevant to the consistency-based approach, which operates on the assumption that LLMs are likely to provide logically inconsistent responses for the same question when they are indecisive and hallucinating facts \cite{zhang2023siren}.
For instance, SelfCheckGPT \cite{manakul2023selfcheckgpt} explored several methods, such as BERTScore \cite{zhang2019bertscore}, to check informational consistency between sampled responses.
\citet{mundler2023self} utilized an additional LLM to detect incorrect facts by checking whether there is a contradiction between two responses given the same context.
Our method shares similarities with these approaches in terms of checking consistency among sampled responses.
Nonetheless, our endorsement scores are calculated at a finer level (fact vs fact).
More importantly, we prioritize improving the quality of final responses after detecting hallucinations.

\section{Conclusion}
In this paper, we present self-endorsement, a framework that alleviates hallucinations and improves reasoning capability solely by the LLM itself.
Particularly, we first perform fine-grained fact-level comparisons among multiple sampled candidates to identify reliable facts. 
Then, we produce the final response by either selecting from candidates or regenerating based on these facts.
We evaluate our approach on popular benchmarks including Biographies for the longform generation, TriviaQA for open-domain question answering, and GSM8K for mathematical multi-step reasoning.
Results show that self-endorsement can significantly benefit small or open-source LLMs without intricate instructions compared with previous approaches.

\section*{Limitations}
The main limitation of self-endorsement lies in the computation cost incurred at the fact verification phase.
The cost escalates dramatically when using more candidates for collecting verified facts.
In this work, we have demonstrated the trade-off between candidate numbers and final performance: limited candidate numbers can still help improve factuality and larger models exhibit less sensitivity to hyperparameter selection.
Future studies can also explore quantization \cite{jacob2018quantization} or distilling knowledge into a smaller model \cite{hinton2015distilling} to improve computational efficiency further.
Another limitation is that our method is fully based on prompting.
Given the sensitivity of LLMs to input prompts, the choice of prompts can impact final performance.
Moreover, a single prompt may not consistently yield optimal results across diverse tasks or models.
Techniques used for prompt searching can help solve this problem \cite{yang2023large}. We leave this as future work.

\bibliographystyle{acl_natbib}
\bibliography{custom}

\appendix

\label{sec:appendix}
\section{Examples of Self-Endorsement Prompts}
\begin{figure*}[t]
    \centering
    \includegraphics[width=0.95\textwidth]{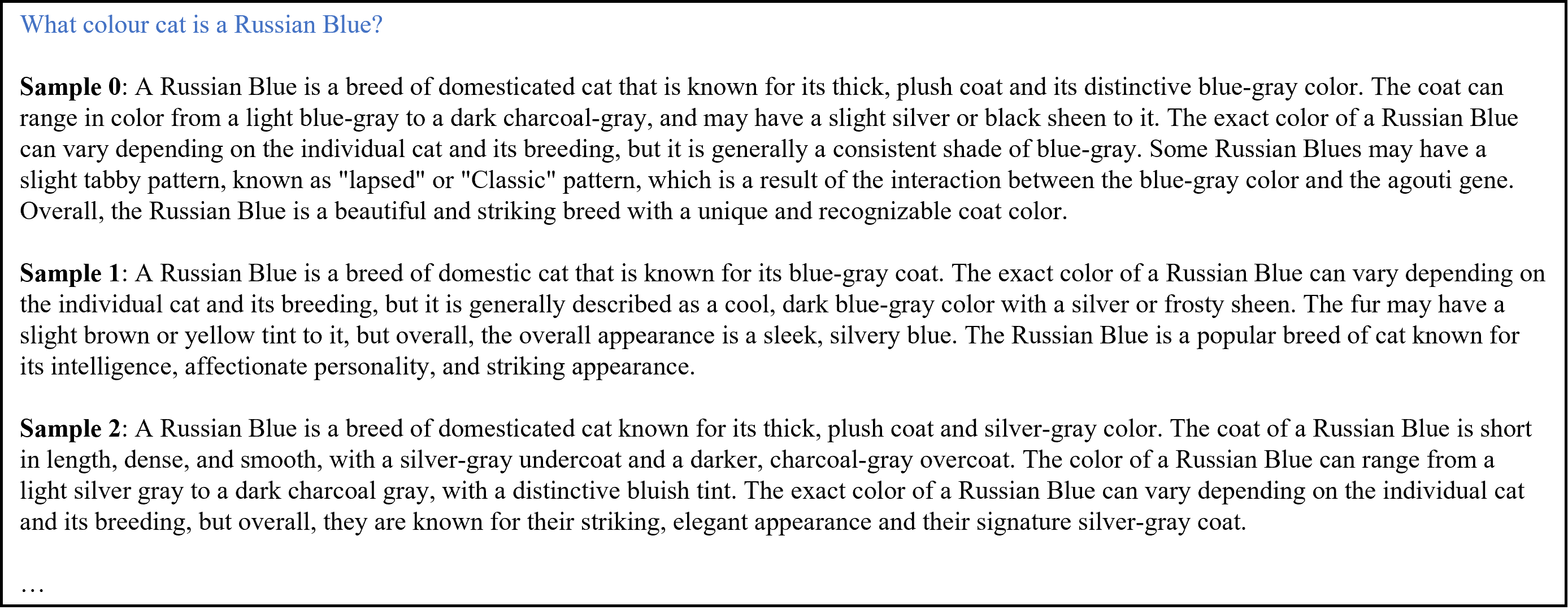}
    \caption{Step 1 - candidate sampling. We only display 3 candidate samples here and the input prompt is highlighted in {\color{blue} blue}.}
    \label{fig:step1}
\end{figure*}

\begin{figure*}[t]
    \centering
    \includegraphics[width=0.95\textwidth]{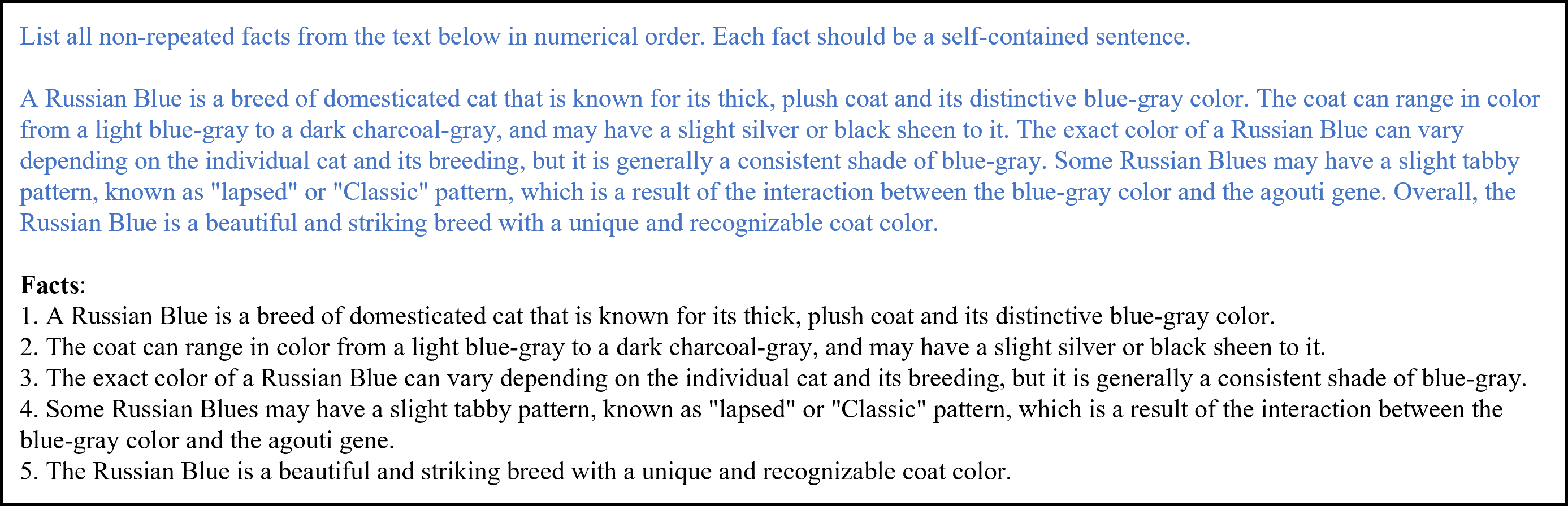}
    \caption{Step 2 - fact decomposition. We take sample 0 in Figure \ref{fig:step1} as an example.}
    \label{fig:step2}
\end{figure*}

\begin{figure*}[t]
    \centering
    \includegraphics[width=0.95\textwidth]{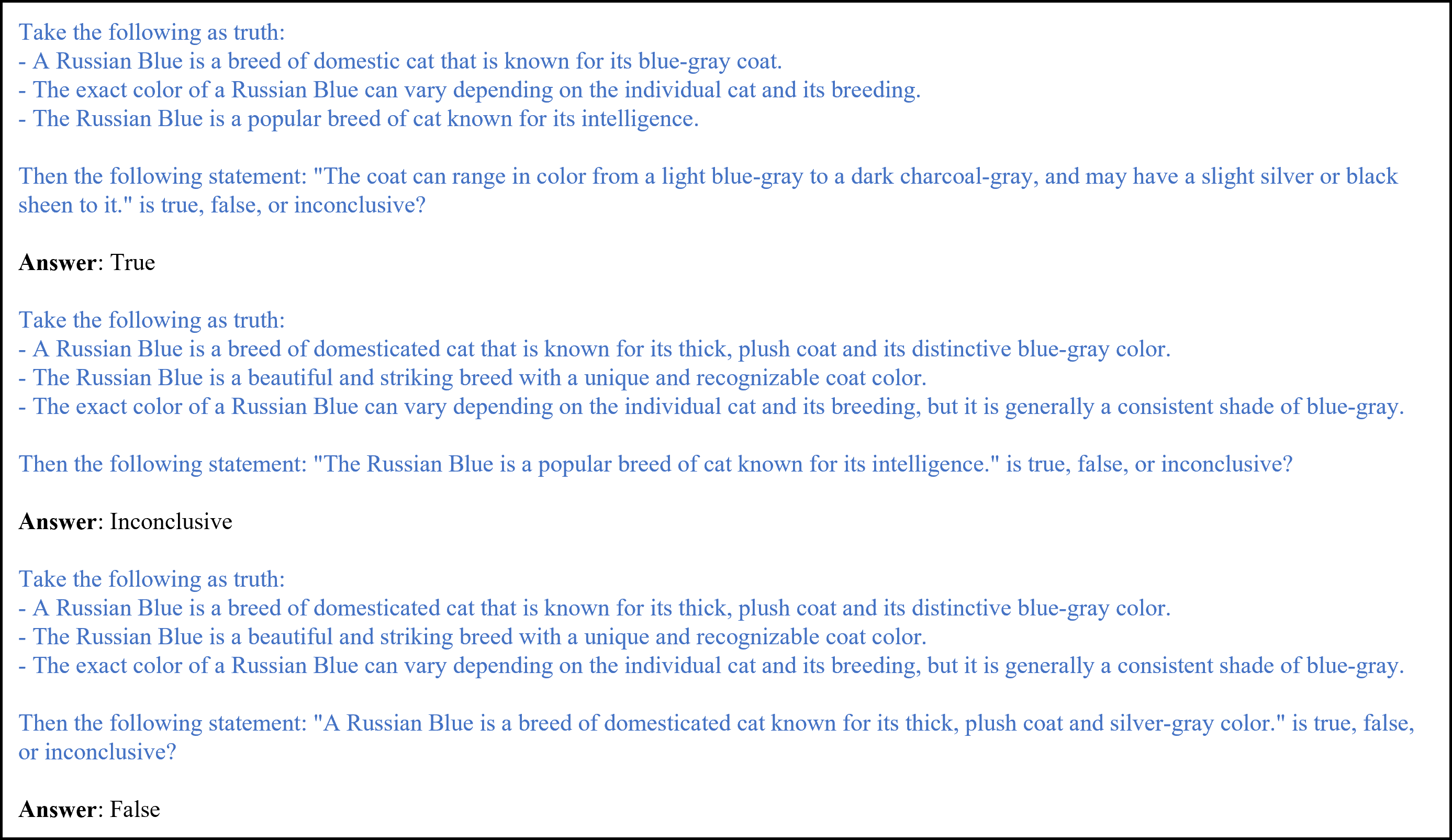}
    \caption{Step 3 - fact verification. We display 3 examples with different classification results.}
    \label{fig:step3}
\end{figure*}

\begin{figure*}[t]
    \centering
    \includegraphics[width=0.95\textwidth]{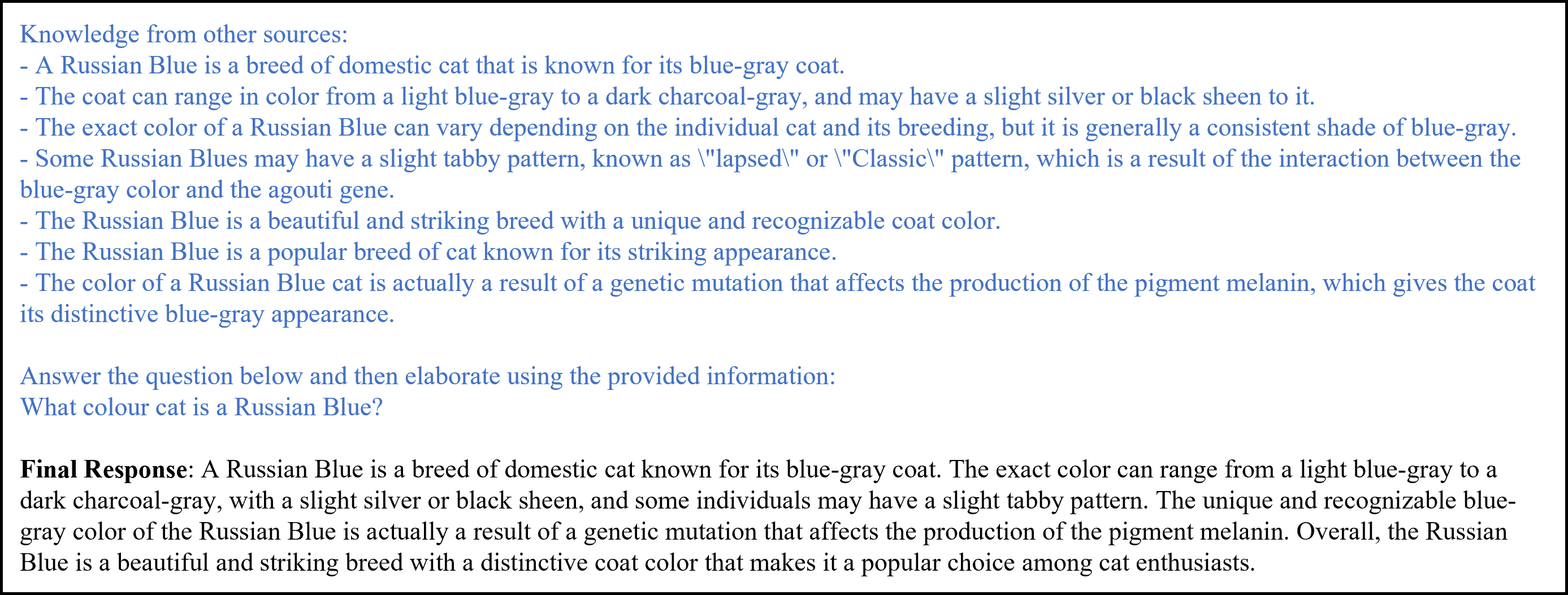}
    \caption{Step 4 - final response production. We present the prompt used for short-text generation.}
    \label{fig:step4}
\end{figure*}

We present an example of our implementation on LLaMA2-70B-Chat, sourced from TriviaQA, as depicted in Figures \ref{fig:step1}, \ref{fig:step2}, \ref{fig:step3}, and \ref{fig:step4}. The responses of this dataset are concise and thus more suitable to display.

\begin{figure*}[t]
    \centering
    \includegraphics[width=0.95\textwidth]{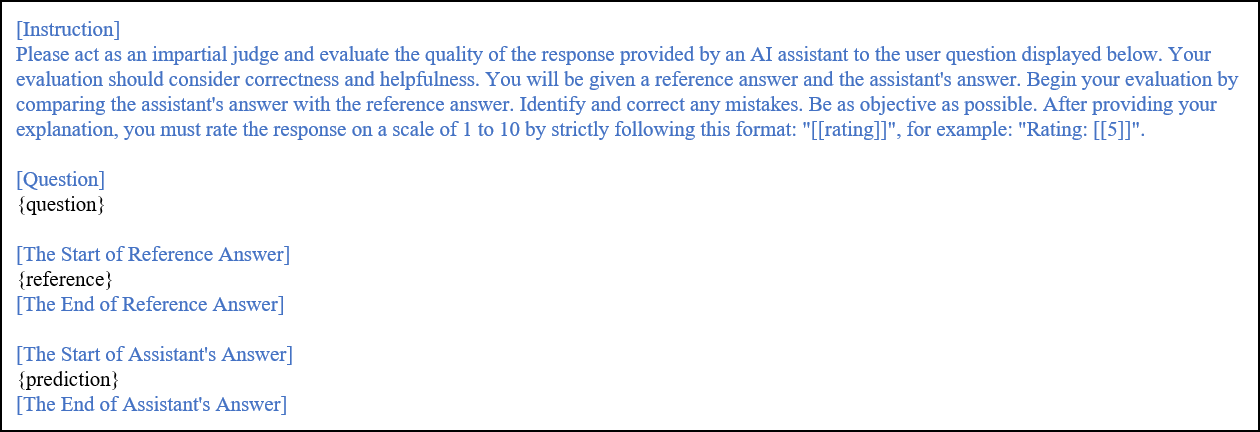}
    \caption{The prompt fed to GPT4 for evaluating model predicted rationales on GSM8K.}
    \label{fig:gpt4_prompt}
\end{figure*}

\end{document}